\documentclass[runningheads]{llncs}
\usepackage[T1]{fontenc}
\usepackage{graphicx}
\usepackage{float}
\usepackage{booktabs} 
\usepackage{graphicx} 
\usepackage{makecell}
\usepackage{multirow}
\usepackage{amssymb}
\usepackage{amsmath}
\usepackage[caption=false]{subfig}
\usepackage{wrapfig}
\usepackage{hyperref}
\usepackage{url}
\usepackage{cite}
\usepackage{bm}
\usepackage{microtype}
\begin{document}
\title{RGB-T Object Detection via Group Shuffled Multi-receptive Attention and Multi-modal Supervision}
\titlerunning{RGB-T Object Detection via GSMA and MS}
%
\author{Jinzhong Wang\inst{1} \and Xuetao Tian\inst{2}  \and Shun Dai\inst{1} \and Tao Zhuo\inst{3} \and Haorui Zeng\inst{1} \and Hongjuan Liu\inst{2} \and Jiaqi Liu\inst{2} \and Xiuwei Zhang\inst{1} \and Yanning Zhang\inst{1}}
\authorrunning{J. Wang et al.}
%
\institute{Northwestern Polytechnical University, Xi’an 710072, China \\
\email{\{xwzhang,ynzhang\}@nwpu.edu.cn}\\ \and
Xi’an ASN Technology Group Co. Ltd, Xi’an 710065, China \\
\email{\{laoda20,liuyixiao198711,a15891296613\}@163.com}\\ \and
Northwest A \& F University, Yangling 712100, China\\
\email{zhuotao724@gmail.com}}
\maketitle              
\begin{abstract}
Multispectral object detection, utilizing both visible (RGB) and thermal infrared (T) modals, has garnered significant attention for its robust performance across diverse weather and lighting conditions. However, effectively exploiting the complementarity between RGB-T modals while maintaining efficiency remains a critical challenge. In this paper, a very simple Group Shuffled Multi-receptive Attention (GSMA) module is proposed to extract and combine multi-scale RGB and thermal features. Then, the extracted multi-modal features are directly integrated with a multi-level path aggregation neck, which significantly improves the fusion effect and efficiency. Meanwhile, multi-modal object detection often adopts union annotations for both modals. This kind of supervision is not sufficient and unfair, since objects observed in one modal may not be seen in the other modal. To solve this issue, Multi-modal Supervision (MS) is proposed to sufficiently supervise RGB-T object detection. Comprehensive experiments on two challenging benchmarks, KAIST and DroneVehicle, demonstrate the proposed model achieves the state-of-the-art accuracy while maintaining competitive efficiency. 

\keywords{Multispectral object detection \and Attention mechanism \and Group shuffle \and Multi-modal supervision.}
\end{abstract}
\section{Introduction}
As an integral branch of computer vision, object detection has a wide range of applications in real-world scenarios. However, unimodal object detection methods often encounter limitations from unfavorable conditions, such as dim lighting, fog, or occlusion~\cite{zheng2018multispectral}. To address this challenge, a common approach is to fuse complementary information of different modals, which has been widely used in tasks such as video surveillance~\cite{alldieck2016context} and autonomous driving~\cite{xiao2020multimodal}. For example, visible cameras typically capture complex details such as color and texture under sufficient lighting. But in dark scenarios, their effectiveness will be significantly reduced. In contrast, thermal cameras specialize in capturing the thermal radiation emitted by objects and are almost unaffected by changes in lighting and weather conditions. Nevertheless, the resolution of thermal images is lower, and the texture and color of objects are absent. Consequently, the sufficient fusion of complementary information of RGB and thermal modals is critical.

Feature-level fusion, also known as middle fusion, has been widely explored since its excellent performance. It often adopts two separated sub-networks to extract feature maps from RGB and thermal modals and employs methods such as channel concatenation~\cite{li2018multispectral} and weighted fusion~\cite{sun2022drone} for further fusion. Researchers also explored more complex fusion modules to fully utilize the potential complementary information between RGB and thermal modal, such as illumination-aware techniques~\cite{guan2019fusion} and attention modules~\cite{zhang2019cross}. However, these methods are typically based on two-stage R-CNN variants~\cite{zhang2019weakly,kim2021uncertainty} and with overly complex designs fail to achieve an optimal balance between accuracy and efficiency. In addition, prevalent studies often utilize union annotations~\cite{zhou2020improving} as detection supervision. It may cause the network easily affected by noise in weak alignment or modal-absent situations. Moreover, using union annotations to supervise two modals is unfair, since objects observed in one modal may not be seen in the other modal. It may cause confusion and failure to fully utilize the precise information of each modal.

In this paper, we propose a novel one-stage SAMS-YOLO network to address the problems mentioned above. Specifically, for significant and efficient multi-modal feature fusion, we introduce a lightweight multi-scale attention module to extract RGB-T multi-receptive field features and combine them via a novel parameter-free group shuffle operation. Through the multi-level path aggregation neck~\cite{liu2018path}, the combined multi-modal features are effectively and sufficiently fused. Additionally, to ensure robust and accurate object detection, we propose a multi-modal supervision strategy consisting of three branches for detection, i.e., RGB, thermal, and fusion, which is supervised by visible, thermal, and union annotations separately. It can solve the problem of unfair supervision caused by union annotations. By integrating the aforementioned lightweight and efficient modules into the one-stage YOLOv5~\cite{Yolov5} framework, we achieve a good balance between detection accuracy and efficiency. Extensive experiments are conducted on the KAIST and DroneVehicle datasets, the results demonstrate superior detection performance. The contributions of this paper are summarized as follows:
\begin{itemize}
    \item [1)] 
    A simple Group Shuffled Multi-receptive Attention (GSMA) module is proposed to effectively extract and combine multi-modal multi-receptive field features. Through the integration with the top-down and bottom-up PANet~\cite{liu2018path}, multi-modal features are efficiently and sufficiently fused. With this module, we achieve a reduction of 2.07\%, 2.28\%, and 1.93\% on $MR^{-2}$ across all-day, day, and night subsets of the KAIST dataset, respectively.
    \item [2)]  
    A Multi-modal Supervision (MS) strategy is proposed to effectively guide the network to learn precise and robust feature representations by leveraging visible, thermal, and union annotations as supervision. This strategy leads to a reduction of 2.66\%, 2.87\%, and 2.25\% on $MR^{-2}$ across all-day, day, and night subsets of the KAIST dataset, respectively.
    \item [3)]
    The proposed RGB-T object detection method, namely SAMS-YOLO, integrates GSMA and MS into YOLOv5, enhancing the detection ability of small targets, night scenes, and occlusion situations. It achieves the state-of-the-art results on two challenging datasets: KAIST multispectral pedestrian dataset and DroneVehicle remote sensing dataset, while maintaining a fast processing speed.
\end{itemize}

\begin{figure*}[t]
    \centering
    \includegraphics[width=\linewidth]{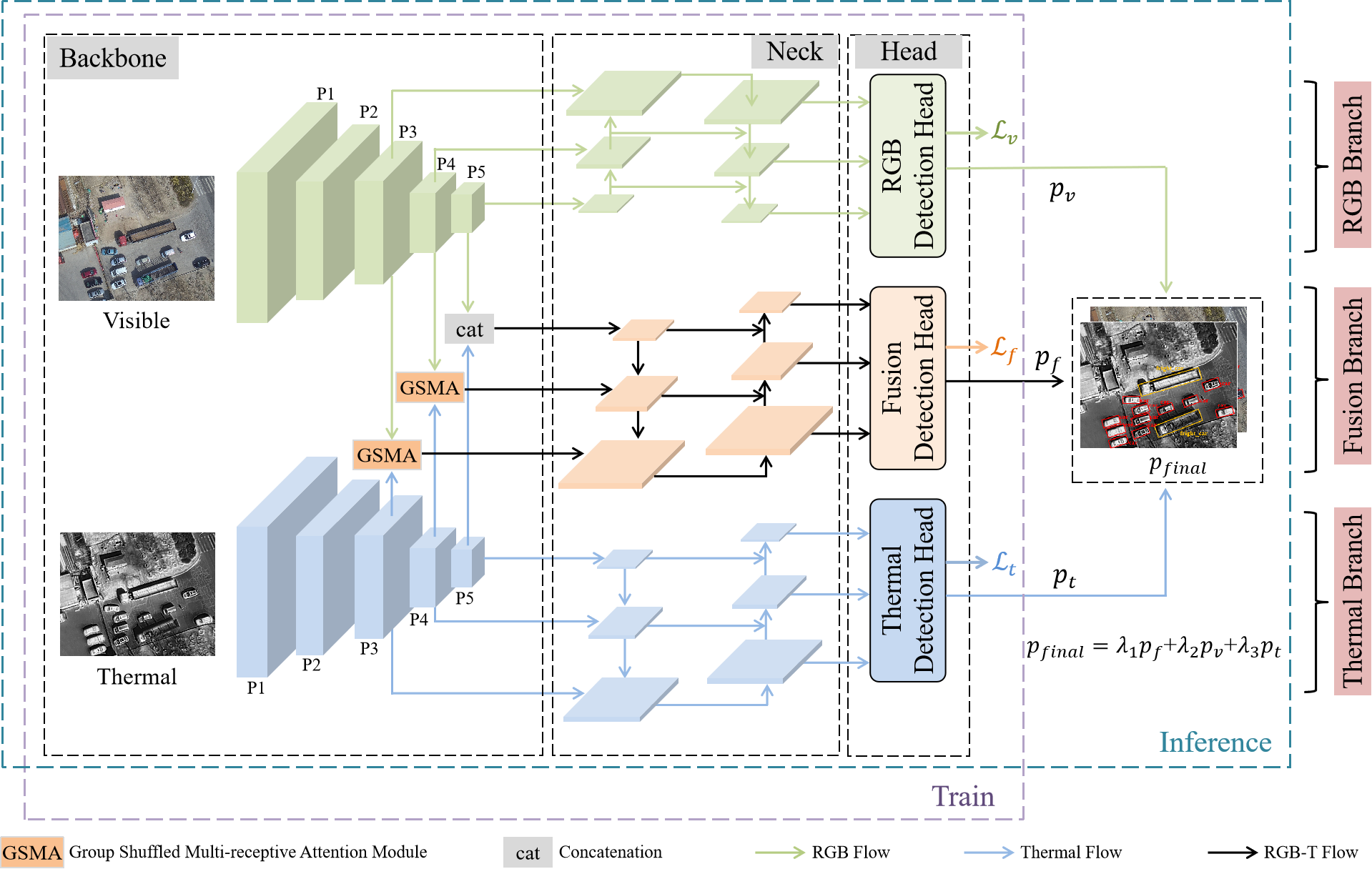}
    \caption{Architecture of the proposed SAMS-YOLO. The multi-modal supervision strategy is applied to the RGB, thermal, and fusion branches. During training, the RGB, thermal, and union annotations are used as supervision to calculate detection loss. During inference, a decision-level fusion is applied to fuse the RGB, thermal, and fusion branch results.}
    \label{fig_framework}
\end{figure*}

\section{Related Work}
\subsection{Multispectral Object Detection}
Due to the significant advantages offered by collaborative detection in visible and thermal domains, multispectral object detection has made remarkable progress. Liu et al.~\cite{liu2016multispectral} adopted a two-stage method Faster R-CNN~\cite{ren2015faster} as the framework, and incorporated two separate pedestrian detectors on visible and thermal images, respectively. SDS-RCNN~\cite{brazil2017illuminating}, MSDS-RCNN~\cite{li2018multispectral}, and I$^{2}$MDet~\cite{zhang2023oriented} leveraged ground truth bounding boxes as weak segmentation annotations to facilitate supervised learning. IAF R-CNN~\cite{li2019illumination} integrated illumination-aware modules into the detection network, enabling dynamic weight adjustment for different input modals based on varying light conditions. AR-CNN~\cite{zhang2019weakly} and TSFADet~\cite{yuan2022translation} proposed modal alignment operations to solve the problem of temporal and spatial misalignment between RGB-T modals. Additionally, Zhou et al.~\cite{zhou2020improving} investigated the issue of multi-modal imbalance resulting from the inadequate fusion of modal information in their MBNet framework. UA-CMDet~\cite{sun2022drone} tackled the quantification of uncertainty associated with multi-modal targets through uncertainty perception and illumination estimation. Li et al.~\cite{10114594} proposed multiscale cross-modal homogeneity enhancement and confidence-aware feature fusion in their MCHE-CF. Notably, most of the approaches mentioned above are based on two-stage R-CNN variants, which suffer from slow detection speeds due to their intricate architecture and multiple stages involved.

\subsection{Multi-modal Features Fusion}
Fusing multi-modal features is a crucial aspect of multispectral object detection, which can be categorized into four types: early fusion, middle fusion, late fusion, and decision-level fusion. Among these, middle fusion strategies have been widely explored and demonstrated to be more effective, as they are more flexible in design and enable deeper feature fusion~\cite{dasgupta2022spatio, qingyun2022cross}. MSDS-RCNN~\cite{li2018multispectral} and UA-CMDet~\cite{sun2022drone} employed a simple channel concatenation approach for feature fusion. CSAA~\cite{cao2023multimodal} combined channel switching and channel concatenation. CIAN~\cite{zhang2019cross} introduced a cross-modal interaction attention module to adaptively recalibrate channel responses. MBNet~\cite{zhou2020improving} leveraged the differences between modals to design a differential modal-aware fusion module. SC-MPD~\cite{dasgupta2022spatio} incorporated a spatial-contextual feature aggregation block to efficiently utilize multiple source features. DCMNet~\cite{xie2022learning} improved feature complementarity through dynamic local and non-local feature aggregation modules. C$^{2}$Former-S$^{2}$ANet~\cite{yuan2023mathbf} employed an intermodality cross-attention module to obtain the calibrated and complementary features between the RGB-T modals. However, a simple concatenation fusion method cannot guarantee accuracy, and complex modules significantly result in high memory usage and latency. To address these challenges, we propose a novel group shuffled multi-receptive attention module that considers both channel and multi-spatial level attention while ensuring low computational costs.

\section{Method}
Fig.~\ref{fig_framework} illustrates the overall architecture of our SAMS-YOLO model. It comprises three primary components: the dual-stream feature extractor backbone, three-branch detection neck, and three-branch detection head. In the training phase, we utilize RGB, thermal, and union annotations of the RGB-T modals for detection supervision. For inference, we adopt a decision-level fusion strategy to weigh the prediction from the RGB, thermal, and fusion branches.

\subsection{Framework Overview}
As shown in Fig.~\ref{fig_framework}, the network takes RGB-T image pairs as input, and a dual-stream backbone with five layers (denoted as P1 to P5) to extract hierarchical feature maps. The size of features generated by P1 to P5 is 2, 4, 8, 16, and 32 times downsampling of the input images, respectively. The neck and head are designed with a three-branch structure of RGB, thermal, and fusion. Firstly, the visible and thermal features obtained by P3 to P5, are passed to the corresponding neck branch, i.e., the top one and the bottom one, for refined feature representation and prediction, respectively. Secondly, two GSMA modules are employed to extract RGB-T multi-receptive field features and fully combine them at the P3 and P4 stages. Thirdly, the concatenated features at the P5 stage and the fused features enhanced by GSMA are then fed into the fusion neck and head (middle branch) for further prediction. The neck here is the top-down and bottom-up PANet~\cite{liu2018path}. Subsequently, as expressed in Eq.~\ref{eq_fusion}, the final prediction result ${p}_{final}$ is obtained by taking the weighted average of fusion detection ${p}_{f}$, visible detection ${p}_{v}$, and thermal detection ${p}_{t}$. $\lambda_{1}$, $\lambda_{2}$, and $\lambda_{3}$ are hyper-parameters, and detailed in the experimental implementation.
\begin{equation}
{p}_{final} = \lambda_{1}{p}_{f} + \lambda_{2}{p}_{v} + \lambda_{3}{p}_{t}  \label{eq_fusion}
\end{equation}

\subsection{Group Shuffled Multi-receptive Attention Module}    
\begin{figure}[t]
    \centering
    \subfloat[GSMA module]{\includegraphics[width=0.55\columnwidth]{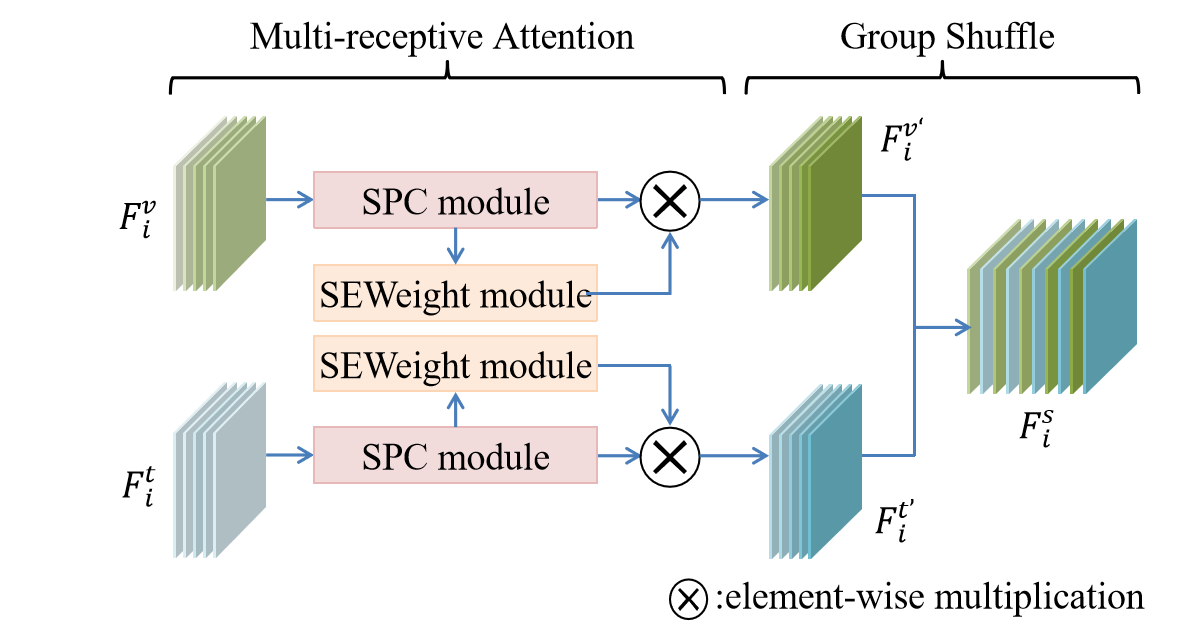} \label{fig_MFFM_arc}}
    \subfloat[SPC module]{\includegraphics[width=0.35\columnwidth]{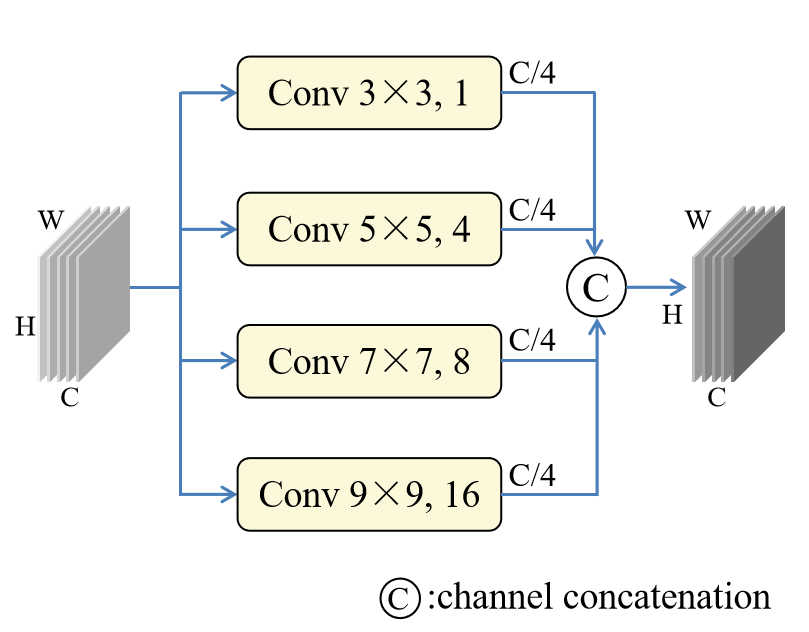} \label{fig_SPC}}
    \caption{The structure of Group Shuffled Multi-receptive Attention module. (a) shows the data flow structure of the GSMA. (b) shows the SPC structure in (a).}
    \label{fig_MFFM}
\end{figure}

The motivation of this work is to build an efficient and effective multi-modal attention mechanism to improve multi-modal feature fusion. As illustrated in Fig.~\ref{fig_MFFM} (a), the structure of the GSMA module is quite straightforward, it contains two parts: multi-receptive attention and group shuffle.

\textbf{Multi-receptive Attention.} Previous studies have rarely focused on the impact of multi-receptive field features on multi-modal feature fusion. Inspired by ~\cite{zhang2022epsanet}, we introduce a multi-receptive attention mechanism to effectively extract the multi-modal multi-scale spatial information. As shown in Fig.~\ref{fig_MFFM} (a), two Squeeze Pyramid Concat (SPC) modules~\cite{zhang2022epsanet} are adopted to obtain multi-receptive field feature maps on channel-wise of the input features $F_{i}^{v}$ and $F_{i}^{t}$ ($i\in \left \{ 3,4 \right \} $). Then, two SEWeight modules~\cite{zhang2022epsanet} are applied to extract channel-wise attention weights for the RGB-T multi-scale features obtained from the SPC module. After that, element-wise multiplication is applied to recalibrate the weights and corresponding feature maps. Finally, the refined features $F_{i}^{v'}$ and $F_{i}^{t'}$ at different receptive fields are combined by the group shuffle operation to obtain $F_{i}^{s}$. 

The structure of the SPC module is shown in Fig.~\ref{fig_MFFM} (b). The input features are extracted by multi-scale group convolution kernels to capture information regarding different spatial resolutions and depths. The multi-scale group convolution kernel sizes are $3\times3$, $5\times5$, $7\times7$, and $9\times9$, with corresponding convolution groups set as 1, 4, 8, and 16. Then, the multi-receptive features are merged through channel concatenation.

\begin{figure}[t]
    \centering
    \includegraphics[width=0.55\columnwidth]{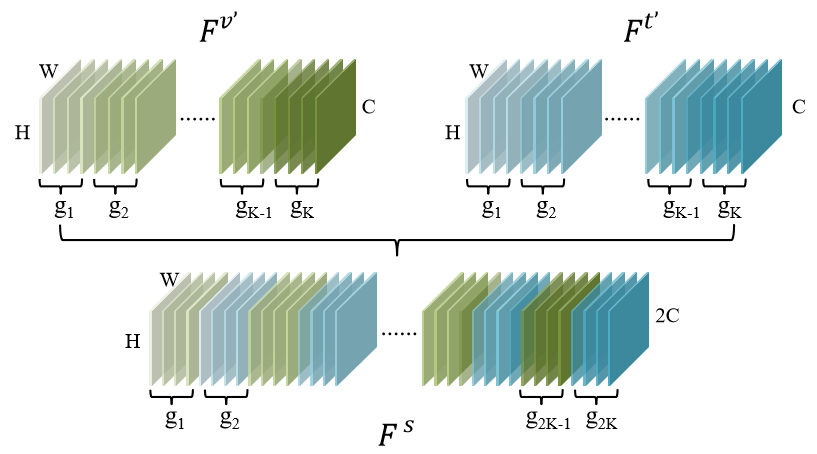}
    \caption{The structure of the Group Shuffle.}
    \label{fig_groupshuffle}
\end{figure}

\textbf{Group Shuffle.} The RGB-T features contain rich complementary information such as color, texture, and contour. To efficiently learn modal correlations, we propose a new representation module called group shuffle. As shown in Fig.~\ref{fig_groupshuffle}, we first split and group the RGB and thermal features along the channel dimension, and then combine them through alternating merging. This parameter-free operation not only preserves the similarity of intra-group modal but also makes inter-group modal responses more diverse, effectively improving the fusion of multi-modal features. Assuming that both RGB and thermal features have C channels, we split these channels into K groups, each containing N channels, where N=C/K. The channel index $j$ of $F_{j}^{v}$ and $F_{j}^{t}$ is mapped to new position ${j}'$ according to Eq.~\ref{eq_shuffle}. It should be noted that when K = 1, group shuffle is equivalent to channel concatenation, and when K = C, it is equivalent to channel shuffle~\cite{zhang2018shufflenet}. 
\begin{equation}
\begin{aligned}     \label{eq_shuffle}
{j}' = 
\begin{cases}
j \bmod N + \left\lfloor \frac{j}{N} \right\rfloor \times 2N, & F_j^{v} \in F^{v} \\
j \bmod N + \left\lfloor \frac{j}{N} \right\rfloor \times 2N + N, & F_j^{t} \in F^{t}
\end{cases}
\end{aligned}
\end{equation}

The multi-modal features after the group shuffle are fully mixed with each other at different multi-receptive fields. By aggregating through multi-level top-down and bottom-up path neck, a comprehensive fusion of multi-modal and multi-scale features can be achieved, and the detection ability of small targets, night scenes, and occlusion situations can be improved.

\subsection{Multi-modal Supervision Strategy}
Due to the possibility of spatial misalignment between visible and thermal images, the position of the same object may be different in the two modals. Simply using the union annotations of RGB-T modals may lead the network to be subjected to biased supervised information due to misalignment, which is not conducive to learning more accurate feature representations. Furthermore, in situations where specific modal are not available, such as a visible camera in dark night environment, the union annotations used to supervise the visible branch may introduce noise disturbance. Therefore, we use visible annotations, thermal annotations, and the union annotations of RGB-T modals to provide more accurate feature learning supervision for feature extraction and detection. As shown in Fig.~\ref{fig_framework}, during training, the visible, thermal, and union annotations are used to supervise the prediction of the RGB, thermal, and fusion branches, respectively.

Additionally, to guide the network to extract more accurate feature representations, following~\cite{li2018multispectral}, we add segmentation prediction heads in the dual-stream backbone and use the RGB and thermal ground truth bounding boxes as segmentation supervision. This method is only used during training and does not affect the network inference speed.

\subsection{Loss Function}
The loss function is built upon YOLOv5~\cite{Yolov5}, incorporating the multi-modal supervision loss and segmentation supervision loss. As shown in Eq.~\ref{eq_loss}, $\mathcal{L}^{f}$, $\mathcal{L}^{v}$ and $\mathcal{L}^{t}$ are fusion, visible and thermal detection loss, respectively. $\mathcal{L}_{cls}$, $\mathcal{L}_{obj}$, $\mathcal{L}_{bbox}$ represents the object classification loss, object confidence loss, and object coordinate position loss, respectively. $\mathcal{L}_{seg}$ is segmentation loss of binary cross-entropy. $\lambda_{cls}$, $\lambda_{obj}$, $\lambda_{bbox}$ and $\lambda_{seg}$ are correction factors and are detailed in the experimental implementation.

\begin{eqnarray}
\resizebox{.9\hsize}{!}{$\begin{aligned}
\mathcal{L}_{total} = & \lambda_{cls} \left ( \mathcal{L}_{cls}^{f} \!+\! \mathcal{L}_{cls}^{v} \!+\! \mathcal{L}_{cls}^{t}  \right )  \!+\! \lambda_{obj}  \left ( \mathcal{L}_{obj}^{f} \!+\! \mathcal{L}_{obj}^{v} \!+\! \mathcal{L}_{obj}^{t} \right ) \!+\! \lambda_{bbox}  \left ( \mathcal{L}_{bbox}^{f} \!+\! \mathcal{L}_{bbox}^{v} \!+\! \mathcal{L}_{bbox}^{t} \right ) \!+\! \lambda_{seg}\mathcal{L}_{seg}
\label{eq_loss}
\end{aligned} $}
\end{eqnarray}

\begin{figure}[t]
    \centering
    \subfloat[Original]{\includegraphics[width=0.48\columnwidth]{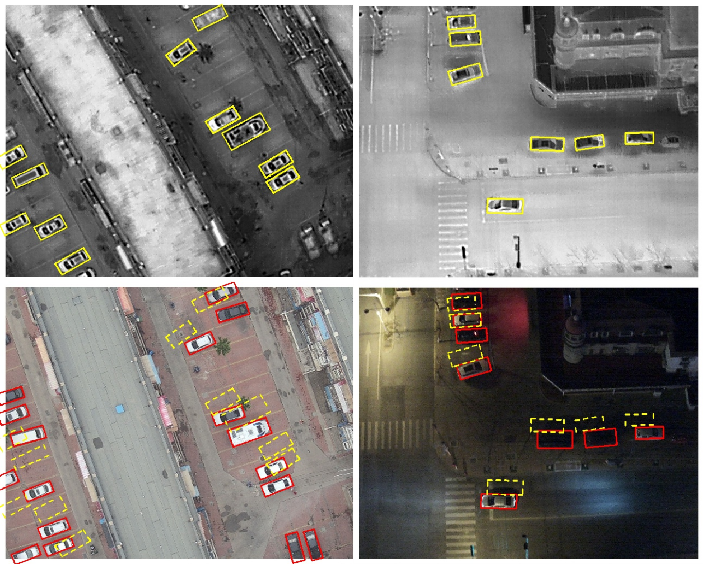} \label{fig_original}}
    \hfill
    \subfloat[Aligned]{\includegraphics[width=0.48\columnwidth]{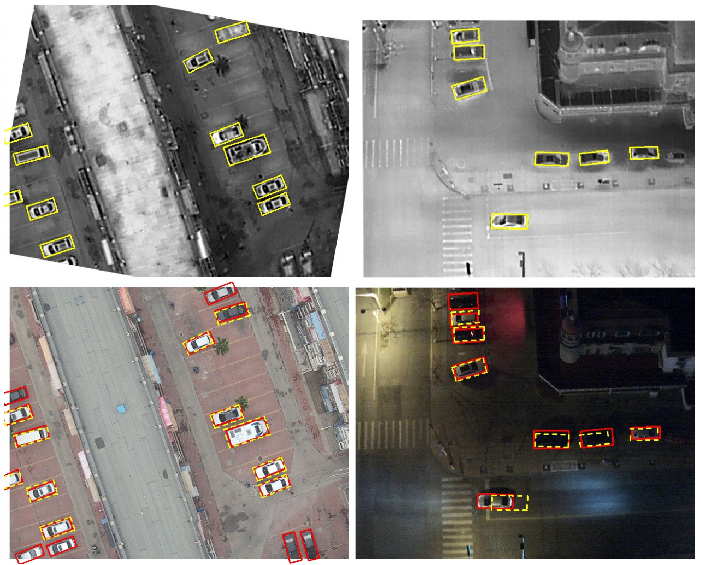} \label{fig_improved}}
    \caption{Illustration of the modal misalignment problem in the DroneVehicle training set. (a) and (b) depict the original and aligned RGB-T image pairs, where the top images are thermal images and the bottom images are visible images. Yellow boxes indicate annotations on the thermal images, while red boxes indicate annotations on the visible images. Both modal annotations are visualized on the visible image.}
    \label{fig_annotation}
\end{figure}

\section{Experiments} 
In this section, the experimental datasets are introduced firstly. Then, the implementation details and comparison experiments are presented. Finally, ablation studies are conducted to verify the effectiveness of each component in our approach.

\subsection{Datasets and Evaluation Metric}
Our approach is evaluated on two challenging datasets: KAIST multispectral pedestrian dataset~\cite{hwang2015multispectral} and DroneVehicle remote sensing dataset~\cite{sun2022drone}. These two datasets along with their related evaluation metrics are depicted as follows.

\textbf{KAIST Dataset.} The KAIST dataset~\cite{hwang2015multispectral} consists of 95,328 RGB-T image pairs with 103,128 pedestrian annotations. Following~\cite{liu2016multispectral,zhang2019weakly}, we use 7,095 image pairs for training and 2,252 image pairs for testing. Specifically, The test images contain 1,455 day-time images (‘Day’) and 797 night-time images (‘Night’). We use the standard $MR^{-2}$ (log-average Miss Rate over false positive per image range of $\left [ 10^{-2},10^{0} \right ]$)~\cite{hwang2015multispectral} and FPS (frames per second) to evaluate the performance. Note that a lower $MR^{-2}$ indicates a better detection performance. 

\textbf{DroneVehicle Dataset.} The DroneVehicle dataset~\cite{sun2022drone} is a large-scale drone-based image dataset of oriented vehicles. It contains 28,439 RGB-T image pairs with 953,087 instances covering urban roads, residential areas, parking lots, and other scenarios. Specifically, it contains five categories, i.e., car, truck, bus, van, and freight car. 

Due to the lack of union annotations in the DroneVehicle training set, to verify the effectiveness of our method, we established union annotations for the first time by taking the union of independent annotations of RGB and thermal modals. Specifically, due to the cross-modal misalignment problems~\cite{yuan2022translation}, we use the method proposed by~\cite{zhang2023learning} to register 2,441 RGB-T image pairs in the training set. As shown in Fig.~\ref{fig_annotation}, the position misalignment problem has been addressed compared with the original image pairs. Finally, we use the aligned training set for training and the original test set for evaluation. Following~\cite{sun2022drone}, we evaluate the detection performance by utilizing the mean average precision (mAP) under different IoU thresholds as the evaluation metric. Specifically, we select mAP$_{0.5}$ and mAP in our experiments. Here, the mAP indicates that the IoU threshold is set from 0.50 to 0.95 with a step of 0.05. Note that the evaluation performance of RGB and thermal modal are averaged as the final evaluation results in our experiment.

\subsection{Implementation Details}
The proposed SAMS-YOLO is based on YOLOv5~\cite{Yolov5}. During training, mosaic data enhancement, random HSV enhancement, and horizontal flip are adopted to enhance RGB-T image pairs. The input images in both datasets are resized to $640 \times 640$ pixels. The optimizer employed is stochastic gradient descent (SGD) for 150 epochs with a learning rate of 0.001 and a batch size of 6. Weight decay and momentum are set to 0.0001 and 0.937, respectively. The hyper-parameters $\lambda_{1}$, $\lambda_{2}$, and $\lambda_{3}$ in Eq.~\ref{eq_fusion} are set to 0.5, 0.25 and 0.25, respectively. The correction factors $\lambda_{cls}$, $\lambda_{obj}$, $\lambda_{bbox}$, and $\lambda_{seg}$ in Eq.~\ref{eq_loss} are set to 0.5, 1.0, 0.05, and 0.25, respectively.
\begin{table}[t]    
    \centering
    \caption{Evaluation results on the KAIST dataset.}
    \label{table_KAIST}
    \resizebox{0.8\linewidth}{!}{
    \begin{tabular}{l|c|c|c|c|c|c}
    \hline
    \multicolumn{2}{c|}{\multirow{2}*{Method}} & \multicolumn{3}{c|}{$MR^{-2}$} & \multirow{2}*{Platform} & \multirow{2}*{FPS} \\\cline{3-5}  
    \multicolumn{2}{c|}{} & All-Day & Day & Night & ~ & ~\\
    \hline
    ACF~\cite{hwang2015multispectral} & RGB+IR & 47.32 & 42.57 & 56.17 & MATLAB & 0.37 \\
    Halfway Fusion~\cite{liu2016multispectral} & RGB+IR & 25.75 & 24.88 & 26.59 & TITAN X & 2.33 \\
    IATDNN + IASS~\cite{guan2019fusion} & RGB+IR & 14.95 & 14.67 & 15.72 & TITAN X & 4.00 \\
    CIAN~\cite{zhang2019cross} & RGB+IR & 14.12 & 14.77 & 11.13 & 1080Ti & 14.29 \\
    MSDS-RCNN~\cite{li2018multispectral} & RGB+IR & 11.34 & 10.53 & 12.94 & TITAN X & 4.55 \\ 
    AR-CNN~\cite{zhang2019weakly} & RGB+IR & 9.34 & 9.94 & 8.38 & 1080Ti & 8.33 \\ 
    CMPD~\cite{li2022confidence} & RGB+IR & 8.16 & 8.77 & 7.31 & 1080Ti & 9.09 \\
    MBNet~\cite{zhou2020improving} & RGB+IR & 8.13 & 8.28 & 7.86 & 1080Ti & 14.29 \\ 
    SC-MPD~\cite{dasgupta2022spatio} & RGB+IR & 8.07 & 8.16 & 7.51 & Tesla P6 & 10 \\ 
    BAANet~\cite{yang2022baanet} & RGB+IR & 7.92 & 8.37 & 6.98 & 1080Ti & 14.29 \\
    UGCML~\cite{kim2021uncertainty} & RGB+IR & 8.18 & 6.96 & 7.89 & 1080Ti & 11.11 \\ 
    CPFM~\cite{10382506} & RGB+IR & 7.09 & 5.61 & 6.62 & 3090Ti & - \\ 
    MCHE-CF~\cite{10114594} & RGB+IR & 6.71 & 7.58 & 5.52 & - & - \\
    DCMNet~\cite{xie2022learning} & RGB+IR & 5.84 & 6.48 & 4.60 & 3090 & 7.14 \\\hline
    \multirow{2}*{YOLOv5~\cite{Yolov5}} 
    & RGB & 18.72 & 13.48 & 28.45 & 2080Ti & 83.33  \\
    & IR & 16.90 & 22.33 & 6.34 & 2080Ti & 83.33  \\\hline
    SAMS-YOLO (ours) & RGB+IR & \textbf{5.26} & \textbf{6.00} & \textbf{3.81} & 2080Ti & \textbf{19.31} \\ 
    \hline
    \end{tabular}}
\end{table}

\subsection{Comparison on the KAIST Dataset}
The performance of SAMS-YOLO on the KAIST Dataset is presented in Table~\ref{table_KAIST}. Our method achieves the best accuracy, reaching 5.26\%, 6.00\%, and 3.81\% $MR^{-2}$ on the reasonable all-day, day, and night subsets, respectively. At the same time, our detector achieves the fastest detection speed of 19.31 FPS on 2080Ti GPU. Compared with YOLOv5~\cite{Yolov5}, our method has achieved significant performance improvement. This indicates that the proposed GSMA module enhances the fusion ability of complementary information of RGB-T modals, and improves the localization accuracy while maintaining high efficiency. Meanwhile, through the proposed MS strategy, the detection results from the RGB, thermal, and fusion branches ensure the model's recall rate, achieving superior consequences.

\begin{table*}[t]  
    \centering
    \caption{Evaluation results on the DroneVehicle dataset.}
    \label{table_DV}
    \resizebox{\linewidth}{!}
    {\begin{tabular}{l|c|c|c|c|c|c|c|c|c}
    \hline
    Method & car & truck & van & bus & freight car & mAP$_{0.5}$ & mAP & Platform & FPS\\ \hline
    UA-CMDet~\cite{sun2022drone} & 87.5 & 60.7 & 38.0 & 87.1 & 46.8 & 64.00 & -  & 3090 & 9.12 \\ 
    Oriented R-CNN~\cite{Xie_2021_ICCV} & 89.9 & 56.6 & 46.9 & 89.6 & 54.4 & 67.52 & 42.60  & - & - \\ 
    RoI Transformer~\cite{ding2019learning} & 90.1 & 60.4 & 52.2 & 89.7 & 58.9 & 70.29 & 43.57 & - & - \\
    CIAN(OBB)~\cite{zhang2019cross} & 89.98 & 62.47 & 49.59 & 88.9 & 60.22 & 70.23 & - & GV100 & \textbf{21.7} \\
    AR-CNN(OBB)~\cite{zhang2019weakly} & 90.08 & 64.82 & 51.51 & 89.38 & 62.12 & 71.58 & - & GV100 & 18.2 \\
    TSFADet~\cite{yuan2022translation} & 89.88 & 67.87 & 53.99 & 89.81 & 63.74 & 73.06 & - & GV100 & 18.6\\
    ViT-B+RVSA~\cite{wang2022advancing} & 89.7 & 52.3 & 44.4 & 88.0 & 51.0 & 65.07 & 42.63  & - & - \\
    C$^{2}$Former-S$^{2}$ANet~\cite{yuan2023mathbf} & 90.2 & 68.3 & 58.5 & 89.8 & 64.4 & 74.2 & - & TITAN V & - \\
    I$^{2}$MDet~\cite{zhang2023oriented} & 96.3 & 73.4 & 58.6 & 93.2 & 65.0 & 77.30 & 46.20 & - & - \\ 
    DTNet-B~\cite{10494734} & 90.2 & 78.1 & 65.7 & 89.2 & \textbf{67.9} & 78.23 & 52.85 & 3090 & - \\     \hline
    SAMS-YOLO-OBB (ours) & \textbf{97.00} & \textbf{79.57} & \textbf{67.50} & \textbf{95.95} & 63.75 & \textbf{80.75} & \textbf{57.13} & 2080Ti & 17.83\\ 
    \hline
    \end{tabular}}
\end{table*}
\begin{table*}[t] 
\centering
    \caption{Effect of K in group shuffle. We tune the group hyperparameter K to $\left \{ 1,2,4,8,16,32,C\right \}$.} 
    \label{table_group_ablation}
    \resizebox{0.55\linewidth}{!}{ 
    \begin{tabular}{c|c|c|c|c|c|c|c}
    \hline
    \multirow{2}*{Subset} & \multicolumn{6}{c}{$MR^{-2}$} \\
    \cline{2-8} 
    ~ & K=1 & K=2 & K=4 & K=8 & K=16 & K=32 & K=C \\
    \hline
    All-day & 7.30 & 8.11 & 7.63 & 6.89 & \textbf{6.48} & 7.70 & 6.77 \\
    Day & 8.33 & 9.40 & 8.92 & 8.77 & \textbf{7.86} & 10.03 & 8.46 \\
    Night & 5.66 & 6.07 & 5.11 & \textbf{3.26} & 3.94 & 4.10 & 3.64 \\
    \hline
    \end{tabular}}
\end{table*}

\subsection{Comparison on the DroneVehicle Dataset}
Since the DroneVehicle is an oriented bounding box detection dataset, to achieve the detection of oriented objects, we modified our model referring to~\cite{yang2020arbitrary}, named SAMS-YOLO-OBB. The experimental results are shown in Table~\ref{table_DV}. Our method significantly outperforms others, achieving the best performance. Specifically, we improved mAP$_{0.5}$ and mAP metrics by 2.52\% and 4.28\%, respectively, compared with the previous best method DTNet-B. Furthermore, we achieved the highest scores in the subcategories of car, truck, van, and bus, with mAP$_{0.5}$ increasing by 0.70\%, 1.47\%, 1.8\%, and 2.75\% compared to the previous state-of-the-art methods. 

\subsection{Ablation Study}
Ablation experiments are conducted on the KAIST dataset to verify the effect of the GSMA and MS modules. We employ a framework without the GSMA module and RGB and thermal branches in the MS strategy as the baseline in the experiment.

\begin{figure*}[t]
    \centering
    \includegraphics[width=\linewidth]{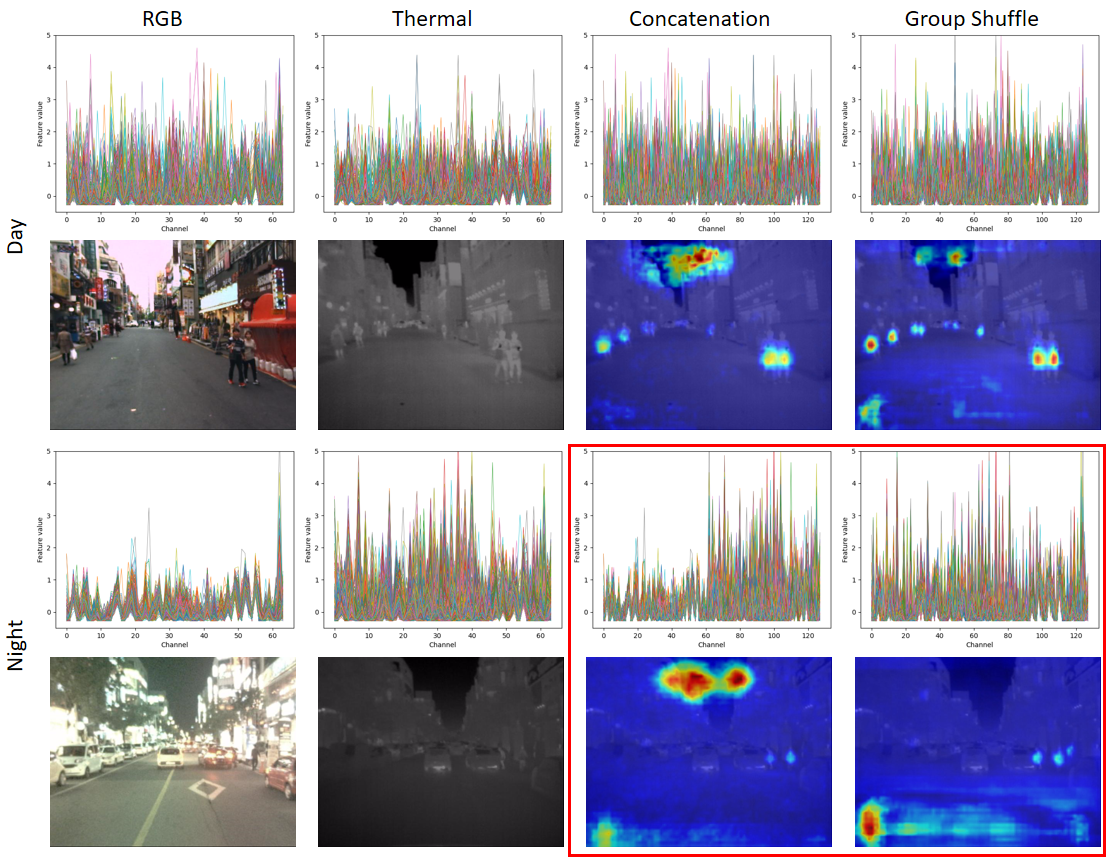}
    \caption{Impacts of RGB-T feature concatenation and group shuffle. The top and third rows depict the feature map values along the channel dimension, while the second and fourth rows display RGB-T images and corresponding heatmaps. The top two rows showcase day-time scenes, whereas the bottom two rows depict night-time scenes. Notably, the response of RGB features diminishes during night-time. As observed in the feature maps and heatmaps in the bottom right corner, compared to simple concatenation, the group shuffle operation achieves more comprehensive multi-modal feature mixing. Through the GSMA module and multi-path aggregation fusion, the network exhibits heightened attention towards pedestrian areas.}
    \label{fig_cat_shf}
\end{figure*}
\begin{table*}[ht!]    
    \centering
    \caption{Effects of GSMA module. MA means multi-receptive attention and GS means group shuffle operation.} \label{table_MFM_ablation}
    \resizebox{\linewidth}{!}{
    \begin{tabular}{c|c|c|c|c|c|c|c|c|c}
    \hline
    \multirow{2}*{Method} & \multicolumn{9}{c}{$MR^{-2}$} \\
    \cline{2-10}
    ~ & All-day & Day & Night & Near & Medium & Far & None & Partial & Heavy \\
    \hline
    Baseline & 8.55 & 10.14 & 5.87 & 0.00 & 13.23 & 41.28 & 22.75 & 28.53 & 50.02 \\
    MA without GS & $7.30_{(-1.25)}$ & $8.33_{(-1.81)}$ & $5.66_{(-0.21)}$ & 0.00 & $12.56_{(-0.67)}$ & $40.63_{-0.65)}$ & $22.13_{(-0.62)}$ & $29.04_{(0.51)}$ & $52.00_{(1.98)}$ \\   
    GS before MA & $7.39_{(-1.16)}$ & $9.78_{(-0.36)}$ & \pmb{$3.90_{(-1.97)}$} & 0.00 & \pmb{$11.46_{(-1.77)}$} & $40.82_{(-0.46)}$ & $21.15_{(-1.60)}$ & $26.83_{(-1.70)}$ & $47.07_{(-2.95)}$ \\
    GS after MA & \pmb{$6.48_{(-2.07)}$} & \pmb{$7.86_{(-2.28)}$} & $3.94_{(-1.93)}$ & 0.00 & $11.67_{(-1.56)}$ & \pmb{$38.51_{(-2.77)}$} & \pmb{$21.00_{(-1.75)}$} & \pmb{$24.86_{(-3.67)}$} & \pmb{$47.03_{(-2.99)}$} \\
    CBAM~\cite{woo2018cbam} & $7.57_{(-0.98)}$ & $9.49_{(-0.65)}$ & $4.45_{(-1.42)}$ & 0.00 & $14.22_{(0.99)}$ & $44.13_{(2.85)}$ & $24.04_{(1.29)}$ & $27.86_{(-0.67)}$ & $54.36_{(4.34)}$ \\
    GCB~\cite{cao2019gcnet} & $8.48_{(-0.07)}$ & $10.58_{(0.44)}$ & $5.07_{(-0.80)}$ & 0.00 & $12.93_{(-0.30)}$ & $39.64_{(-1.64)}$ & $21.86_{(-0.89)}$ & $29.84_{(1.31)}$ & $52.11_{(2.09)}$ \\
    \hline
    \end{tabular}}
\end{table*}

\begin{figure*}[t]
    \centering
    \includegraphics[width=\linewidth]{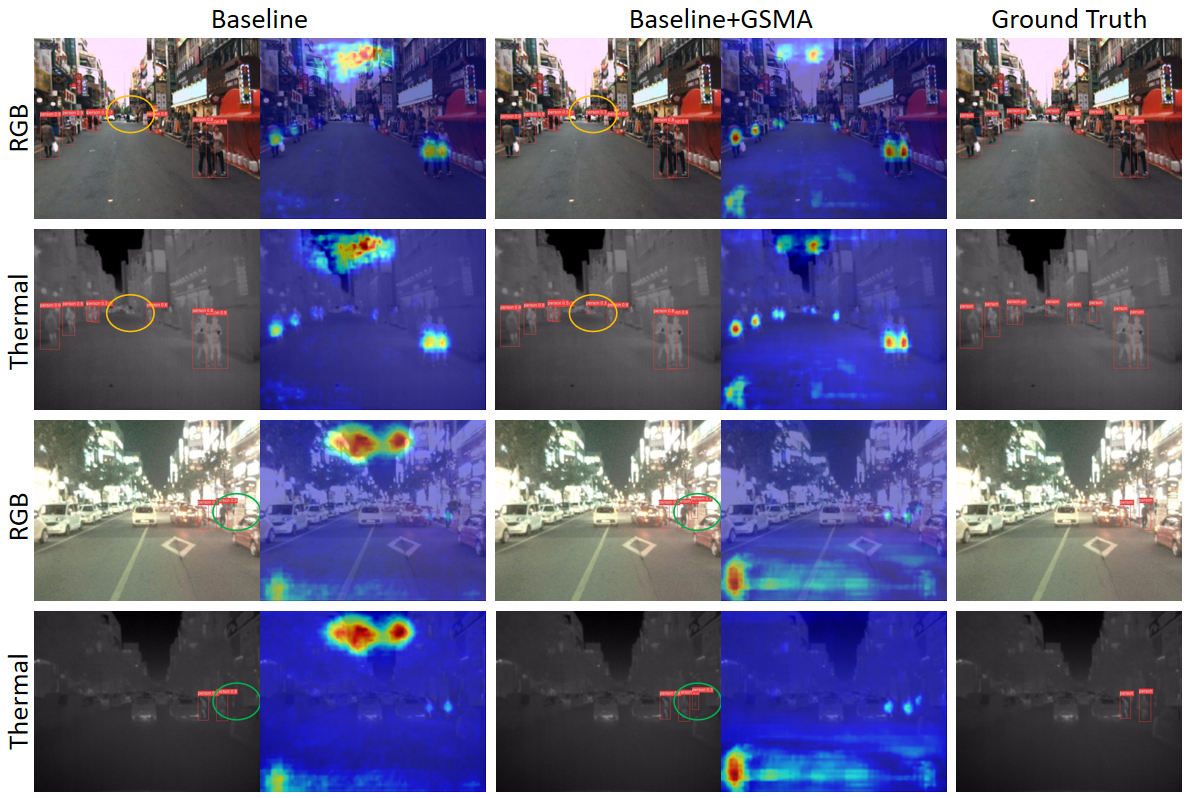}
    \caption{Examples of detection and heatmaps of baseline method and the addition of GSMA module on the KAIST pedestrian dataset. As shown in the orange and green elliptical areas, the GSMA module enhances the detection ability of small and occluded objects.}
    \label{fig_GSMA_ablation}
\end{figure*}
\begin{table*}[ht!]
\centering      
    \centering
    \caption{Effects of GSMA module and MS strategy evaluated on KAIST dataset.}
    \label{table_ablation}
    \resizebox{\linewidth}{!}{
    \begin{tabular}{c|c|c|c|c|c|c|c|c|c|c}
    \hline
    \multirow{2}*{GSMA} & \multirow{2}*{MS} & \multicolumn{9}{c}{$MR^{-2}$} \\ 
    \cline{3-11}
    ~  & ~ & All-day & Day & Night & Near & Medium & Far & None & Partial & Heavy\\ 
    \hline
    × & × & 8.55 & 10.14 & 5.87 & 0.00 & 13.23 & 41.28 & 22.75 & 28.53 & 50.02 \\
    \checkmark & × & $6.48_{(-2.07)}$ & $7.86_{(-2.28)}$ & $3.94_{(-1.93)}$ & 0.00 & $11.67_{(-1.56)}$ & $38.51_{(-2.77)}$ & $21.00_{(-1.75)}$ & $24.86_{(-3.67)}$ & $47.03_{(-2.99)}$ \\
    × & \checkmark & $5.89_{(-2.66)}$ & $7.27_{(-2.87)}$ & $3.62_{(-2.25)}$ & 0.00 & \pmb{$9.21_{(-4.02)}$} & $34.18_{(-7.10)}$ & \pmb{$17.70_{(-5.05)}$} & $23.64_{(-4.89)}$ & $47.13_{(-2.89)}$ \\ 
    \checkmark & \checkmark & \pmb{$5.26_{(-3.29)}$} & \pmb{$6.00_{(-4.14)}$} & $3.81_{(-2.06)}$ & 0.00 & $9.91_{(-3.32)}$ & $36.25_{(-5.03)}$ & $19.05_{(-3.70)}$ & \pmb{$23.04_{(-5.49)}$} & \pmb{$46.87_{(-3.15)}$} \\
    \hline
    \end{tabular}}
\end{table*}

\textbf{Effectiveness of GSMA.} We first conduct experiments on the effects of different group configurations and the operation position of group shuffle. As illustrated in Table~\ref{table_group_ablation}, optimal performance is observed when K=16. As shown in Table~\ref{table_MFM_ablation}, placing group shuffle (GS) after multi-receptive attention (MA) achieves the best performance. Besides, the proposed GSMA module exhibits prominent superiority, when compared to existing attention methods such as CBAM~\cite{woo2018cbam} and GCB~\cite{cao2019gcnet}. As shown in Table~\ref{table_ablation}, when adding the GSMA into the baseline, it achieves the reductions of 2.07\%, 2.28\%, and 1.93\% on $MR^{-2}$ across the reasonable all-day, day, and night subset, respectively. Fig.~\ref{fig_cat_shf} exhibits some typical images and the corresponding visualized feature maps, we can see that through the group shuffle operation, the fused RGB-T features is inclined to highlight pedestrian regions. Relevant detection result examples are shown in Fig.~\ref{fig_GSMA_ablation}. This indicates that the GSMA facilitates a complementary fusion of RGB-T features and enhances detection accuracy in night scenes and situations involving occlusion. 

\textbf{Effectiveness of MS Strategy.} As indicated in Table~\ref{table_ablation}, after incorporating the MS strategy into the baseline, we observe decreases of 2.66\%, 2.87\%, and 2.25\% on $MR^{-2}$ across reasonable all-day, day, and night conditions, respectively. These results indicate that the supervision by utilizing independent annotations for RGB, thermal, and fusion modal is more sufficient and can fully utilize the
precise information of each modal. 

The combination of the GSMA module and MS strategy also verifies their effectiveness. Finally, the baseline was reduced by 3.29\%, 4.14\%, and 2.06\% on the reasonable all-day, day, and night subsets via applying the GSMA module and MS strategy which obtained the best performance. 

\section{Conclusions}  
In this paper, we propose a novel multispectral object detection network named SAMS-YOLO, which can effectively improve multi-modal detection accuracy while maintaining high efficiency. Particularly, we design a group shuffled multi-receptive attention module to fully extract and combine multi-scale RGB-T features and promote deeper multi-modal feature fusion. In addition, we propose a multi-modal supervision strategy to guide the network in learning more accurate and robust feature representations, as well as improving object detection. Comprehensive comparison and ablation experiments on KAIST and DroneVehicle datasets demonstrate the effectiveness of the proposed framework and its components. The proposed method can be applied to unmanned driving, video surveillance, and other RGB-T object detection domains.

%
%
%
\bibliographystyle{splncs04}
\bibliography{SAMS-YOLO}

\end{document}